\documentclass[conference]{IEEEtran}
\IEEEoverridecommandlockouts
\usepackage{cite}
\usepackage{amsmath,amssymb,amsfonts}
\usepackage{graphicx}
\usepackage{textcomp}

\usepackage{booktabs} % For formal tables
\usepackage{arydshln}
\usepackage{subcaption,verbatim,engord,enumitem,multirow,bm,caption}
\usepackage{algorithm,algorithmic}
\usepackage{amscd}
\usepackage{mathtools}
\usepackage{balance}

\newtheorem{definition}{Definition}
\newtheorem{proposition}{Proposition}
\newtheorem{theorem}{Theorem}
\newtheorem{proof}{Proof}

\usepackage{xcolor}
\def\BibTeX{{\rm B\kern-.05em{\sc i\kern-.025em b}\kern-.08em
    T\kern-.1667em\lower.7ex\hbox{E}\kern-.125emX}}
\begin{document}

\title{Scaling Up Collaborative Filtering Data Sets\\
through Randomized Fractal Expansions
}

\author{\IEEEauthorblockN{Francois Belletti}
\IEEEauthorblockA{\textit{Google AI} \\
belletti@google.com}
\and
\IEEEauthorblockN{Karthik Lakshmanan}
\IEEEauthorblockA{\textit{Google AI} \\
lakshmanan@google.com
}
\and
\IEEEauthorblockN{Walid Krichene}
\IEEEauthorblockA{\textit{Google AI} \\
walidk@google.com
}
\and
\IEEEauthorblockN{Nicolas Mayoraz}
\IEEEauthorblockA{\textit{Google AI} \\
nmayoraz@google.com
}
\and
\IEEEauthorblockN{Yi-Fan Chen}
\IEEEauthorblockA{\textit{Google AI} \\
yifanchen@google.com
}
\and
\IEEEauthorblockN{John Anderson}
\IEEEauthorblockA{\textit{Google AI} \\
janders@google.com
}
\and
\IEEEauthorblockN{Taylor Robie}
\IEEEauthorblockA{\textit{Google AI} \\
taylorrobie@google.com
}
\and
\IEEEauthorblockN{Tayo Oguntebi}
\IEEEauthorblockA{\textit{Google AI} \\
tayo@google.com
}
\and
\IEEEauthorblockN{Dan Shirron}
\IEEEauthorblockA{\textit{Intel Nervana} \\
dan.shirron@intel.com
}
\and
\IEEEauthorblockN{Amit Bleiwess}
\IEEEauthorblockA{\textit{Intel Nervana} \\
amit.bleiweiss@intel.com
}
}

\maketitle

\begin{abstract}
Recommender system research suffers from a disconnect between the size of academic data sets and the scale of industrial production systems.
In order to bridge that gap, we propose to generate large-scale user/item interaction data sets by expanding pre-existing public data sets.
Our key contribution is a technique that expands user/item incidence matrices matrices to large numbers of rows (users), columns (items), and non-zero values (interactions). The proposed method adapts Kronecker Graph Theory to preserve key higher order statistical properties such as the fat-tailed distribution of user engagements, item popularity, and singular value spectra of user/item interaction matrices.
Preserving such properties is key to building large realistic synthetic data sets which in turn can be employed reliably to benchmark recommender systems and the systems employed to train them. We further apply our stochastic expansion algorithm to the binarized MovieLens 20M data set, which comprises 20M interactions between 27K movies and 138K users.
The resulting expanded data set has 1.2B ratings, 2.2M users, and 855K items, which can be scaled up or down. \end{abstract}


\section{Introduction}
Machine Learning (ML) benchmarks compare the capabilities of models, distributed training systems and linear algebra accelerators on realistic problems at scale.
For these benchmarks to be effective, results need to be reproducible by many different groups which implies that publicly shared data sets need to be available. 

Unfortunately, while recommendation systems constitute a key industrial application of ML at scale, large public data sets recording user/item interactions on online platforms are not yet available.
For instance, although the Netflix data set \cite{bennett2007netflix} and the MovieLens data set \cite{harper2016movielens} are publicly available, they are orders of magnitude smaller than proprietary data~\cite{covington2016deep,belletti2018factorized,zhao2018categorical}.

\begin{table}[t!]
\center
\caption{Size of MovieLens 20M~\cite{harper2016movielens} vs industrial dataset in~\cite{zhao2018categorical}.}\label{tb:dataset}
\setlength{\tabcolsep}{2.5pt}
\begin{tabular}{lcc}
\toprule
 & \textbf{MovieLens 20M} & \textbf{Industrial}\\
\midrule
\#users  & 138K & Hundreds of Millions\\
\midrule
\#items & 27K & 2M\\
\midrule
\#topics & 19 & 600K\\
\midrule
\#observations & 20M & Hundreds of Billions\\
\bottomrule
\end{tabular}
\end{table}

\textbf{Proprietary data sets and privacy:}
While releasing large anonymized proprietary recommendation data sets may seem an acceptable solution from a technical standpoint, it is a non-trivial problem to preserve user privacy while still maintaining useful characteristics of the dataset.
For instance,\cite{narayanan2006break} shows a privacy breach of the Netflix prize dataset. More importantly, publishing anonymized industrial data sets runs counter to user expectations that their data may only be used in a restricted manner to improve the quality of their experience on the platform.

Therefore, we decide not to make user data more broadly available to preserve the privacy of users. We instead choose to produce synthetic yet realistic data sets whose scale is commensurate with that of our production problems while only consuming already publicly available data.

\textbf{Producing a realistic binary MovieLens 10 billion+ dataset:}
In this work, we focus on the MovieLens dataset which only entails movie ratings posted publicly by users of the MovieLens platform.
The MovieLens data set has now become a standard benchmark for academic research in recommender systems.
Many recent research articles rely on MovieLens~\cite{tang2018caser,krichene2018efficient,he2017neural,liang2016factorization,tu2018structural,zheng2016neural,zhou2018deep,rudolph2016exponential,zhang2017enabling,abdollahpouri2017controlling,lu2016sparse,bhargava2017active,jawanpuria2018unified,nimishakavi2018dual}. The latest version of MovieLens~\cite{harper2016movielens} has accrued more than $800$ citations according to Google Scholar. A binarized version of this dataset is obtained when all the ratings are substituted by $1.0$ (proposed in Neural Collaborative Filtering~\cite{he2017neural}).
While in previous work we have considered the original MovieLens data set comprising ratings on a discrete scale~\cite{belletti2019scalable},
we now focus on its binarized version.
Although the binarized version is representative of industrial collaborative filtering aiming at predicting which item a given user is most likely to view~\cite{covington2016deep}, the data set still only entails few observed interactions and more importantly a very small catalogue of users/items, compared to industrial proprietary recommendation data.

Industrial recommender systems typically have to nominate items from catalogues comprising several million distinct elements.
The large number of observations collected by online platforms about user/item interactions also enables performance gains by increasing the dimension of the embeddings employed to represent items.
In most modern ML recommendations, the model learns a vector valued representer in $\mathbb{R}^d$ for each of the $I$ users/items of the catalog.
Typically each user and item is represented with $d \sim 1e^2 \dots 1e^3$ scalars and $I \sim 1e^6 \dots 1e^9$ elements are present in user set and in the item set.
Storing, accessing and training such vast embedding tables presents unique challenges as large tables will no longer easily fit in the memory of a single machine:
distributed embedding tables are often necessary to store the learned embedding tables;
hierarchical embedding access strategies such as hierarchical softmax~\cite{zhao2018categorical}
or differentiated softmax~\cite{grave2017efficient} provide better data structures and learning paradigms; an appropriate negative sampling strategy~\cite{covington2016deep} or regularization~\cite{krichene2018efficient} is needed to solve the extreme classification problem selecting one item from the catalog constitutents.
By scaling up the public MovieLens data set, we want to move the problem into a regime where such issues are critical so that the corresponding benchmark is helpful for industrial applications.

\begin{figure}
    \centering
    \includegraphics[width=0.5\textwidth]{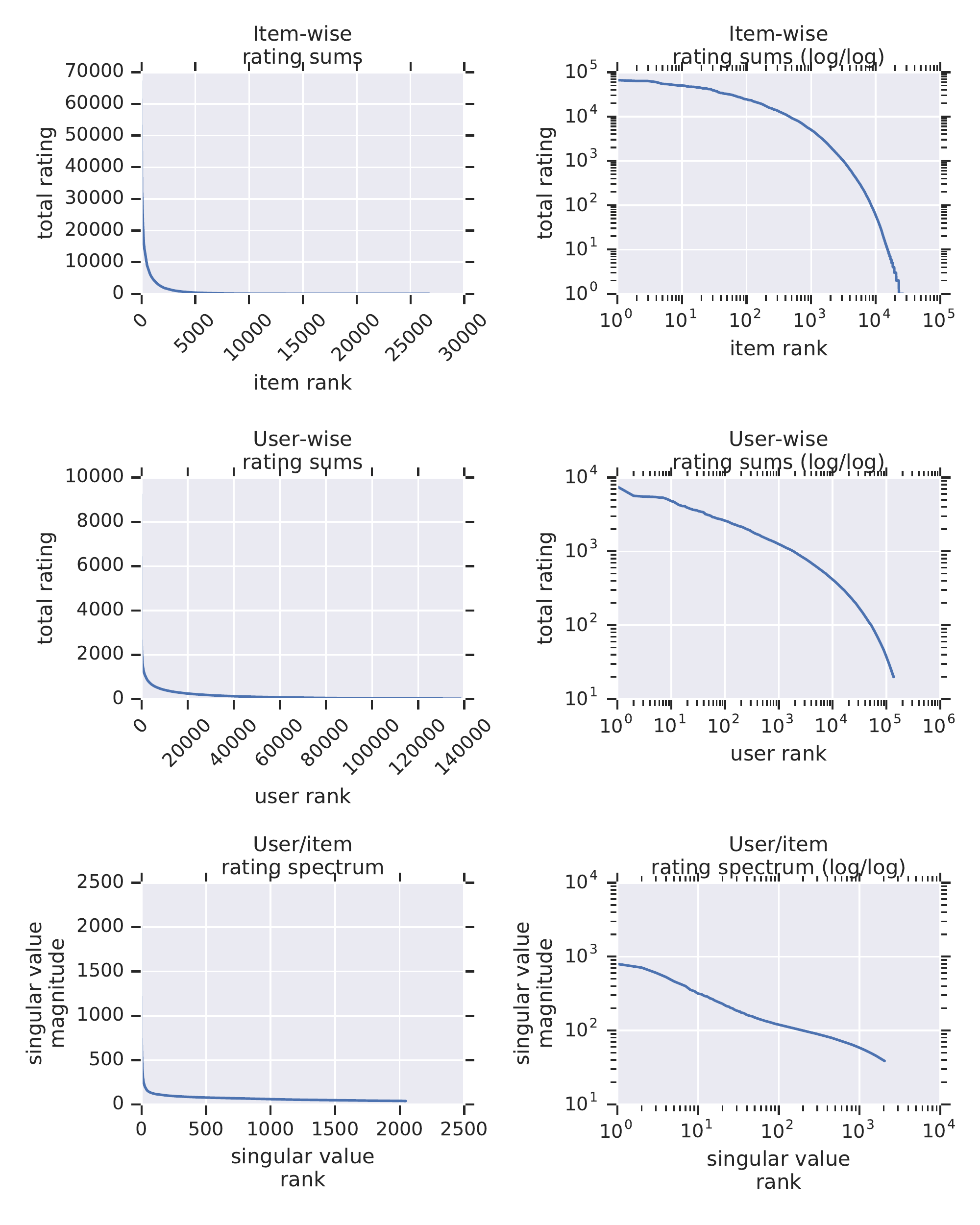}
        \caption{Key first and second order properties of the binarized MovieLens 20m user/item rating matrix we aim to preserve while synthetically expanding the data set. \textbf{Top:} item popularity distribution (total ratings of each item). \textbf{Middle:} user engagement distribution (total ratings of each user). \textbf{Bottom:} dominant singular values of the rating matrix (core to the difficulty of matrix factorization tasks).}
    \label{fig:original_properties}
\end{figure}

In order to provide a new data set --- more aligned with the needs of production scale recommender systems ---
we therefore aim at expanding publicly available data by creating a realistic surrogate.
The following constraints help create a production-size synthetic recommendation problem similar and at least as hard an ML problem as the original one for matrix factorization approaches to recommendations~\cite{koren2009matrix,he2017neural}:
\begin{itemize}
    \item orders of magnitude more users and items are present in the synthetic dataset;
    \item the synthetic dataset is realistic in that its first and second order statistics match those of the original dataset presented in Figure~\ref{fig:original_properties}.
\end{itemize}

Key first and second order statistics of interest we aim to preserve
are summarized in Figure~\ref{fig:original_properties} --- the details of their computation are given in Section~\ref{sec:theory}.

\textbf{Adapting Kronecker Graph expansions to binarized user/item interactions:}
We employ the Kronecker Graph Theory introduced in~\cite{leskovec2005realistic} to achieve a suitable fractal expansion of recommendation data to benchmark linear and non-linear user/item factorization approaches for recommendations~\cite{koren2009matrix,he2017neural}.
Consider a recommendation problem comprising $m$ users and $n$ items.
Let $(R_{i,j})_{i=1 \dots m,j=1 \dots n}$ be the sparse matrix of binarized recorded interactions (i.e. $1$ if user $i$ has consumed item $j$ and $0$ otherwise).
The key insight we develop in the present paper is that a carefully crafted fractal expansion of $R$ can preserve high level statistics of the original data set while scaling its size up by multiple orders of magnitudes.

Many different transforms can be applied to the matrix $R$ which can be considered a standard sparse binary 2 dimensional image. 
A recent approach to creating synthetic recommendation data sets consists in making parametric assumptions on user behavior by instantiating a user model interacting with an online platform~\cite{chaney2017algorithmic,schmit2017human}.
Unfortunately, such methods (even calibrated to reproduce empirical facts in actual data sets) do not provide strong guarantees that the resulting interaction data is similar to the original. A challenging problem in this domain is to build user models that can provide such guarantees, which can be validated using online experiments.
In this work, instead of simulating recommendations in a parametric user-centric way as in~\cite{chaney2017algorithmic,schmit2017human}, we choose a non-parametric approach operating directly in the space of user/item affinity.
In order to synthesize a large realistic dataset in a principled manner, we adapt the Kronecker expansions which have previously been employed to produce large realistic graphs in~\cite{leskovec2005realistic}.
We employ a non-parametric randomized simulation of the evolution of the user/item bi-partite graph to create a large synthetic data set.
It is noteworthy that as opposed to our original approach in~\cite{belletti2019scalable} --- where we put emphasis on analytic tractability --- we now employ a method that loses some analytic tractability but still preserves key statistics of the data set.
Furthermore, we show how randomized operations help address limitations of the previous method which yielded an interaction matrix with a discernible block-wise repetitive structure.

While Kronecker Graphs Theory is developed in~\cite{leskovec2005realistic,leskovec2010kronecker} on square adjacency matrices, the Kronecker product operator is well defined on rectangular matrices and therefore we can apply a similar technique to user/item interaction data sets --- which was already noted in~\cite{leskovec2010kronecker} but not developed extensively.
As in~\cite{leskovec2010kronecker} we will use a stochastic version of the Kronecker extension for a binary original matrix.
The Kronecker Graph generation paradigm has to be changed with the present data set in other aspects. However, we need to decrease the expansion rate to generate data sets with the scale we desire, not orders of magnitude too large. We need to do so while maintaining key conservation properties of the original algorithm~\cite{leskovec2010kronecker}.
Furthermore, we introduce a new block-wise shuffling to randomize the Kronecker operator and yield a data set more helpful to train ML models for recommendations.

In order to reliably employ Kronecker based fractal expansions on recommender system data we devise the following contributions:
\begin{itemize}
    \item we develop a new technique based on linear algebra to adapt fractal Kronecker expansions to recommendation problems;
    \item we introduce a randomly shuffled extension of the original Kronecker product to prevent block-wise structural repetitions and take steps to prevent test data from leaking into the data set employed to train collaborative filtering models;
    \item we also show that the resulting algorithm we develop is scalable and easily parallelizable as we employ it on the actual MovieLens 20 million dataset;
    \item we produce a synthetic yet realistic MovieLens 1.2 billion dataset to help recommender system research scale up in computational benchmark for model training;
    \item we demonstrate that key recommendation system specific properties of the original dataset are preserved by the deterministic version of our technique;
    \item we make the corresponding open source code available so that other researchers may reproduce our findings and tailor the generated synthetic data to their needs.
\end{itemize}

The present article is organized as follows: First, we describe prior research on ML for recommendations and large synthetic dataset generation. Next, we develop a randomized adaptation of Kronecker Graphs to user/item interaction matrices and prove key theoretical properties. Finally, we employ the resulting algorithm experimentally to MovieLens 20m data to validate its statistical properties.

\section{Related work}
Recommender systems constitute the workhorse of many e-commerce, social networking and entertainment platforms.
In the present paper we focus on the classical setting where the key role of a recommender system is to suggest relevant items to a given user.
Although other approaches are very popular such as content based recommendations~\cite{ricci2015recommender} or social recommendations~\cite{chaney2015probabilistic}, collaborative filtering remains a prevalent approach to the recommendation problem~\cite{linden2003amazon,sarwar2001item,resnick1994grouplens}.

\textbf{Collaborative filtering:}
The key insight behind collaborative filtering is to learn affinities between users and items based on previously collected user/item interaction data.
Collaborative filtering exists in different flavors.
Neighborhood methods group users by inter-user similarity and will recommend items to a given user that have been consumed by neighbors~\cite{ricci2015recommender}.
Latent methods try to decompose user/item affinity as the result of the interaction of a few underlying representative factors characterizing the user and the item (e.g. Principal Component Analysis~\cite{jolliffe2011principal}, Latent Dirichlet Allocation~\cite{blei2003latent}). Matrix factorization~\cite{koren2009matrix} is a Latent Factor Method that relies on solving the matrix completion problem to recommend items for users. 

The matrix factorization approach represents the affinity $a_{i,j}$ between a user $i$ and an item $j$ with an inner product $x^T_i y_j$ where $x_i$ and $y_j$ are two vectors in $\mathbb{R}^d$ representing the user and the item respectively.
Given a sparse matrix of user/item interactions $R=(r_{i,j})_{i = 1 \dots m, j=1 \dots n}$, user and item factors can therefore be learned by approximating $R$ with a low rank matrix $X Y^T$ where $X \in \mathbb{R}^{m, k}$ entails the user factors and $Y \in \mathbb{R}^{n, k}$ contains the item factors.
The data set $R$ represents ratings as in the MovieLens dataset~\cite{harper2016movielens} or item consumption ($r_{i,j}=1$ if and only if the user $i$ has consumed item $j$~\cite{bennett2007netflix}) --- the latter being considered here.
The matrix factorization approach is an example of a solution to the rating matrix completion problem which aims at predicting the rating of an item $j$ by a user $i$ which has not been observed yet and corresponds to a value of $0$ in the sparse original rating matrix.
Such a factorization method learns an approximation of the data that preserves a few higher order properties of the rating matrix $R$. 
In particular, the low rank approximation tries to mimic the singular value spectrum of the original data set.
We draw inspiration from matrix factorization to tackle synthetic data generation.
The present paper will adopt a similar approach to extend collaborative filtering data-sets. 
Besides trying to preserve the spectral properties of the original data,
we operate under the constraint of conserving its first and second order statistical properties.

\textbf{Deep Learning for recommender systems:}
Collaborative filtering has known many recent developments which motivate our objective of expanding public data sets in a realistic manner.
Deep Neural Networks (DNNs) are now becoming common in both non-linear matrix factorization tasks~\cite{wang2015collaborative,he2017neural,covington2016deep} and sequential recommendations~\cite{zhou2004intelligent,shani2005mdp,hariri2012context}.
%The latter rely typically on convolutional neural networks, attention networks or recurrent neural networks to map a sequence of temporally ordered user/item interactions to a set of related items~\cite{tang2018caser,li2017neural,wu2017rnn,beutel2018latent}.
The mapping between user/item pairs and ratings is generally learned by training the neural model to predict user behavior on a large data set of previously observed user/item interactions.

DNNs consume large quantities of data and are computationally expensive to train, therefore they give rise to commonly shared benchmarks aimed at speeding up training.
For training, a Stochastic Gradient Descent method is employed~\cite{lecun2015deep}
which requires forward model computation and back-propagation to be run on many mini-batches of (user, item, score) examples.
The matrix completion task still consists in predicting a rating for the interaction of user $i$ and item $j$ although $(i,j)$ has not been observed in the original data-set.
The model is typically run on billions of examples as the training procedure iterates over the training data set.

\textbf{Freshness in recommender systems:}Model freshness is generally critical to industrial recommendations~\cite{covington2016deep} which implies that only limited time is available to re-train the model on newly available data.
The throughput of the trainer is therefore crucial to providing more engaging recommendation experiences and presenting more novel items.
Unfortunately, public recommendation data sets are too small to provide training-time-to-accuracy benchmarks that can be realistically employed for industrial applications.
Too few different examples are available in MovieLens 20m for instance and the number of different available items is orders of magnitude too small. 
In many industrial settings, millions of items (e.g. products, videos, songs) have to be taken into account by recommendation models.
The recommendation model learns an embedding matrices of size $(N, d)$ where $d \sim 10 - 10^3$
and $N \sim 10^6 - 10^9$ are typical values.
As a consequence, the memory footprint of this matrix may dominate that of the rest of the model by several orders of magnitude.
During training, the latency and bandwidth of the access to such embedding matrices have a prominent influence on the final throughput in examples/second.
Such computational difficulties associated with learning large embedding matrices are worthwhile solving in benchmarks.
A higher throughput enables training models with more examples which enables better statistical regularization and architectural expressiveness.
The multi-billion interaction size of the data set used for training is also a major factor that affects modeling choices and infrastructure development in the industry.

\section{Fractal expansions of user/item interaction data sets}
The present section delineates the insights orienting our design decisions when expanding public recommendation data sets.

\subsubsection{Self-similarity in user/item interactions}

\begin{figure}
    \centering
    \includegraphics[width=0.5\textwidth,trim={1cm 5cm 6cm 0}]{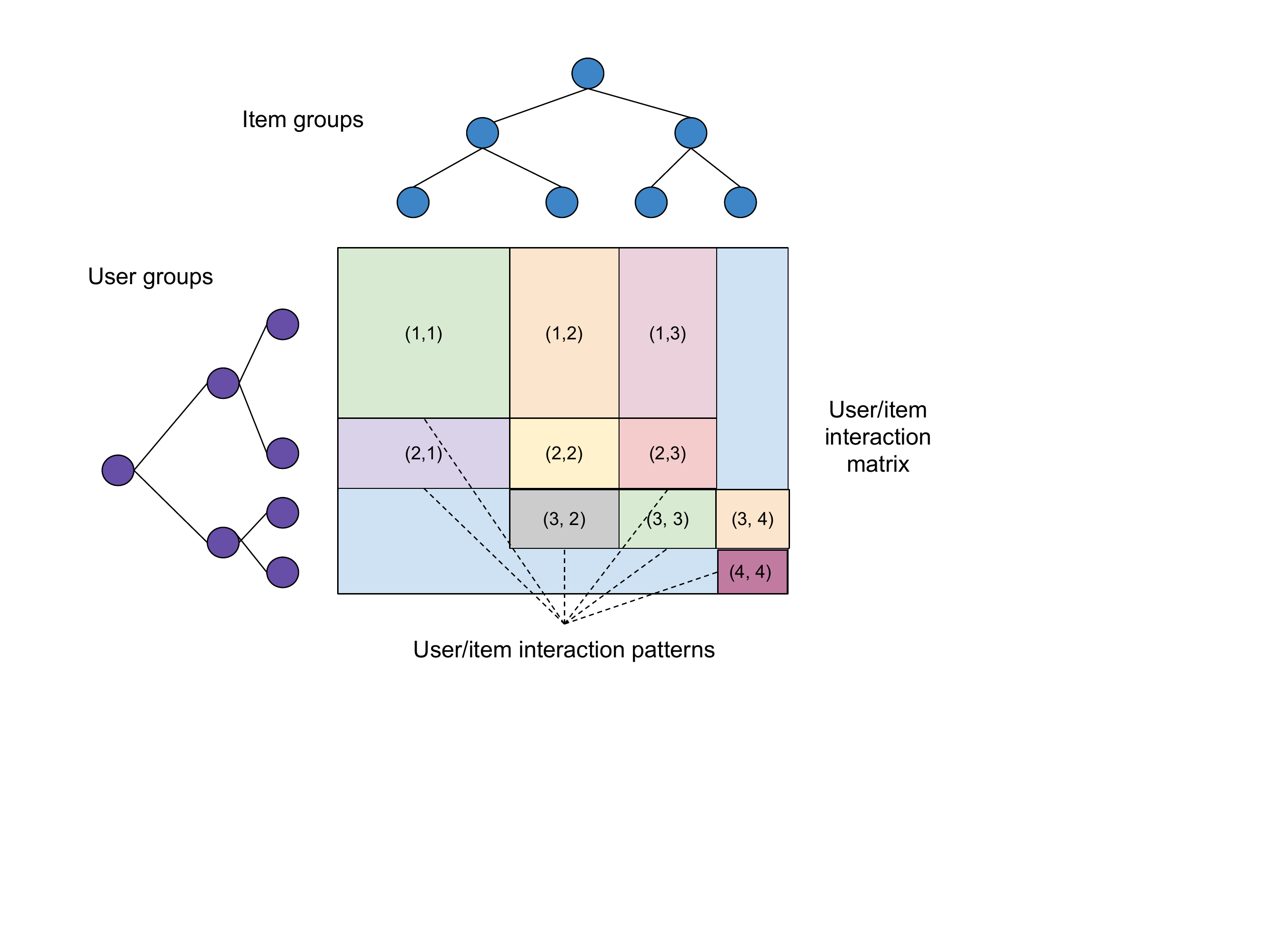}
    \caption{Typical user/item interaction patterns in recommendation data sets. Self-similarity appears as a natural key feature of the hierarchical organization of users and items into groups of various granularity.}
    \label{fig:user_item_patterns}
\end{figure}

Interactions between users and items follow a natural hierarchy in data sets where items can be organized in topics, genres, and categories~\cite{zhao2018categorical}.
There is for instance an item-level fractal structure in MovieLens 20m with a tree-like structure of genres, sub-genres, and directors.
If users were clustered according to their demographics and tastes, another hierarchy would be formed~\cite{ricci2015recommender}.
The corresponding structured user/item interaction matrix is illustrated in Figure~\ref{fig:user_item_patterns}.
The hierarchical nature of user/item interactions (topical and demographic) makes the recommendation data set structurally self-similar (i.e. patterns that occur at more granular scales resemble those affecting coarser scales~\cite{mandelbrot1982fractal}).

One can therefore build a user-group/item-category incidence matrix with user-groups as rows and item-categories as columns
--- a coarse interaction matrix.
As each user group consists of many individuals and each item category comprises multiple movies, the original individual level user/item interaction matrix may be considered as an expanded version of the coarse interaction matrix.
We choose to expand the user/item interaction matrix by extrapolating this self-similar structure and simulating its growth to yet another level of granularity:
original items and users are considered fictional topic and user groups in the expanded data set.

A key advantage of this fractal procedure is that it may be entirely non-parametric and designed to preserve high level properties of the original dataset.
In particular, a fractal expansion re-introduces the patterns originally observed in the entire real dataset within each block of local interactions of the synthetic user/item matrix.
By carefully designing the way such blocks are produced and laid out, we can therefore hope to produce a realistic yet much larger rating matrix.
In the following, we show how the Kronecker operator enables such a construction.

\subsubsection{Fractal expansion through Kronecker products}

The Kronecker product --- denoted $\otimes$ --- is a non-standard matrix operator with an intrinsic self-similar structure:
\begin{equation}
\label{eq:kronecker_product}
A \otimes B =
\begin{bmatrix}
    a_{11} B & \dots & a_{1n} B \\
    \vdots & \ddots & \vdots \\
    a_{m1} B & \dots & a_{mn} B
\end{bmatrix}
\end{equation}
where $A \in \mathbb{R}^{m,n}$, $B \in \mathbb{R}^{p,q}$ and $A \otimes B \in \mathbb{R}^{mp,nq}$.

In the original presentation of Kronecker Graph Theory~\cite{leskovec2005realistic} as well as the stochastic extension~\cite{mahdian2007stochastic} and the extended theory~\cite{leskovec2010kronecker}, the Kronecker product is the core operator enabling the synthesis of graphs with exponentially growing adjacency matrices.
As in the present work, the insight underlying the use of Kronecker Graph Theory in~\cite{leskovec2010kronecker} is to produce large synthetic yet realistic graphs.
The fractal nature of the Kronecker operator as it is applied multiple times (see Figure 2 in~\cite{leskovec2010kronecker} for an illustration) fits the self-similar statistical properties of real world graphs such as the internet, the web or social networks~\cite{leskovec2005realistic}.

If $A$ is the adjacency matrix of the original graph, fractal expansions are created in~\cite{leskovec2005realistic} by chaining Kronecker products as follows:
$$
    A \otimes A \dots A.
$$
As adjacency matrices are square, Kronecker Graphs are not employed on rectangular matrices in pre-existing work although the operation is well defined.
We show that these differences do not prevent Kronecker products from preserving core properties of binarized rating matrices.
A more important challenge is the size of the original matrix we deal with: $R \in \mathbb{R}^{(138 \times 10^3, 27 \times 10^3)}$.
A naive Kronecker expansion would therefore synthesize a rating matrix with $19$ billion users which is too large.

Thus, although Kronecker products seem like an ideal candidate for the mechanism at the core of the self-similar synthesis of a larger recommendation dataset, some modifications are needed to the algorithms developed in~\cite{leskovec2010kronecker}.

\subsubsection{Reduced Kronecker expansions}
We choose to synthesize a user/item rating matrix 
$$
\widetilde{R}
=
\widehat{R}
\otimes
R
$$
where
$
\widehat{R}
$
is a matrix derived from $R$ but much smaller (for instance $\widehat{R} \in \mathbb{R}^{128, 256}$).
For reasons that will become apparent as we explore some theoretical properties of Kronecker fractal expansions, we want to construct a smaller derived matrix $\widehat{R}$ that shares similarities with $R$. In particular, we seek $\widehat{R}$ with a similar row-wise sum distribution (user engagement distribution), column-wise distribution (item engagement distribution) and singular value spectrum (signal to noise ratio distribution in the matrix factorization).

\subsubsection{Implementation at scale and algorithmic extensions}
Computing a Kronecker product between two matrices $A$ and $B$ is an inherently parallel operation.
It is sufficient to broadcast $B$ to each element $(i, j)$ of $A$ and then multiply $B$ by $a_{i,j}$.
Such a property implies that scaling can be achieved.
Another advantage of the operator is that even a single machine can produce a large output data-set by sequentially iterating on the values of $A$. Only storage space commensurable with the size of the original matrix is needed to compute each block of the Kronecker product.
It is noteworthy that generalized fractal expansions can be defined by altering the standard Kronecker product. 
We consider such extensions here as candidates to engineer more challenging synthetic data sets. 
One drawback though is that these extensions may not preserve analytic tractability.

A first generalization defines a binary operator $\otimes_{F}$ with 
$
F
:
\mathbb{R} \times \mathbb{R}^{m,n} \times \mathbb{N}
\rightarrow
\mathbb{R}^{p, q}
$
as follows:
\begin{equation}
\label{eq:gen_kron_product}
A
\otimes_F
B= 
\begin{bmatrix}
    F(a_{11}, B, \omega_{11}) & \dots & F(a_{1n}, B, \omega_{1n}) \\
    \vdots & \ddots & \vdots \\
    F(a_{m1}, B, \omega_{m1}) & \dots & F(a_{mn}, B, \omega_{mn})
\end{bmatrix}    
\end{equation}

where $\omega_{11}, \dots, \omega_{mn}$ is a sequence of pseudo-random numbers.
Including randomization and non-linearity in $F$ appears as a simple way to synthesize data sets entailing more varied patterns. The algorithm we employ to compute Kronecker products is presented in Algorithm~\ref{alg:kron_exp}. The implementation we employ is trivially parallelizable. We only create a list of Kronecker blocks to dump entire rows (users) of the output matrix to file. This is not necessary and can be removed to enable as many processes to run simultaneously and independently as there are elements in $\widehat{R}$ (provided pseudo random numbers are generated in parallel in an appropriate manner).

\begin{algorithm}
\caption{Kronecker fractal expansion $\odot_{F}$}
\label{alg:kron_exp}
\begin{algorithmic}
\FOR{$i = 1$ to $m$}
    \STATE{kBlocks $\gets $ empty list}
    \FOR{$j = 1$ to $n$}
        \STATE{$\omega$ $\gets$ next pseudo random number}
        \STATE{kBlock $\gets$ $F(\widehat{R}(i, j), R, w)$}
        \STATE{kBlocks append kBlock}
    \ENDFOR
    \STATE{outputToFile(kBlocks)}
\ENDFOR
\end{algorithmic}
\end{algorithm}

The only reason why we need a reduced version $\widehat{R}$ of $R$ is to control the size of the expansion.
Also, $A \otimes B$ and $B \otimes A$ are equal after a row-wise and a column-wise permutation.
Therefore, another family of appropriate extensions may be obtained by considering
\begin{equation}
\label{eq:random_sk_kron}
B \otimes_G A =
\begin{bmatrix}
    b_{11} G(A, \omega_{11}) & \dots & b_{1n} G(A, \omega_{1n}) \\
    \vdots & \ddots & \vdots \\
    b_{m1} G(A, \omega_{m1}) & \dots & b_{mn} G(A, \omega_{mn})
\end{bmatrix}
\end{equation}
where 
$
G
:
\mathbb{R}^{m,n} \times \mathbb{N}
\rightarrow
\mathbb{R}^{m',n'}
$
is a randomized sketching operation on the matrix $A$ which reduces its size by several orders of magnitude.
A trivial scheme consists in sampling a small number of rows and columns from $A$ at random. Other random projections~\cite{achlioptas2003database,li2006very,fradkin2003experiments} may of course be used.
The randomized procedures above produce a user/item interaction matrix where there is no longer a block-wise repetitive structure. Less obvious statistical patterns can give rise to more challenging synthetic large-scale collaborative filtering problems.

\subsubsection{Stochastic Kronecker product and dropout for binary rating matrices}
As opposed to our original approach~\cite{belletti2019scalable} which focused on item ratings, we now consider an original binary rating $R$.
In such a setting, a standard Kronecker product is not suitable as the ratings all take the same value of $1$ and therefore multiplications by elements of $\widehat{R}$ do not produce ratings that are all still binary.
We instead use the stochastic Kronecker graph approach from~\cite{leskovec2010kronecker} and employ the reduced matrix's elements $\widehat{R}$ as dropout rates over the matrix $R$.
When computing the block $i,j$ of the expanded rating matrix, instead of using $\widehat{R}_{i,j} R$, we instead consider 
$\text{dropout}(R, rate=\widehat{R}_{i,j})$ after having re-scaled $\widehat{R}$ so that all its elements are in $[0, 1]$.
For each element $k,l$ of $R$, the dropout function for a rate $\widehat{R}_{i,j}$ samples independently from a Bernoulli distribution with parameter $\widehat{R}_{i,j}$. If the sampled number is $1$ $R_{k,l}$, is kept unchanged, otherwise it is dropped and set to $0$.
Such a $block$ dropout operator enjoys statistical properties that are similar to the Kronecker product~\cite{leskovec2010kronecker} while being readily employable on binary data-sets.
The stochastic Kronecker product we devise can therefore be written as follows in matrix notation:
\begin{align*}
&A
\otimes_\text{rand}
B=\\
&\begin{bmatrix}
    \text{Sh}\left(\text{drop}(1 - a_{11}, B) \right) & \dots & \text{Sh}\left(\text{drop}(1 - a_{1n}, B)\right) \\
    \vdots & \ddots & \vdots \\
    \text{Sh}\left(\text{drop}(1 - a_{m1}, B) \right) & \dots & \text{Sh}\left(\text{drop}(1 - a_{mn}, B)\right)
\end{bmatrix}    
\end{align*}
where ``Sh'' denotes the random row-wise and column-wise shuffling operator and ``drop'' denotes the dropout operator whose first argument is the dropout rate and whose second argument is the matrix from which to zero out elements at random.
Algorithm~\ref{alg:kron_randomized_exp} exposes the implementation of the randomized Kronecker product $\odot_{\text{rand}}$.

\subsubsection{Randomized shuffling and Kronecker SVD}
Another limitation of the synthetic data set initially presented in~\cite{belletti2019scalable} is the block-wise repetitive structure of Kronecker products.
Although the synthetic data set is still hard to factorize as the product of two low rank matrices because its singular values are still distributed similarly to the original data set, it is now easy to factorize with a Kronecker SVD~\cite{kamm2000optimal} which takes advantage of the block-wise repetitions in Eq~(\ref{eq:kronecker_product}).
Randomized fractal expansions which presented in Eq~(\ref{eq:gen_kron_product}) and Eq~(\ref{eq:random_sk_kron}) address this issue. The approach we adopt consists in shuffling rows and columns of each block in Eq (\ref{eq:kronecker_product}) independently at random. The shuffles will break the block-wise repetitive structure and prevent Kronecker SVD from producing a trivial solution to the factorization problem.

As a result, the expansion technique we present appears as a reliable first candidate to train linear matrix factorization models~\cite{ricci2015recommender} and non-linear user/item similarity scoring models~\cite{he2017neural}.

\begin{algorithm}
\caption{Kronecker fractal expansion for binary data sets with random shuffling and dropout $\odot_{\text{rand}}$}
\label{alg:kron_randomized_exp}
\begin{algorithmic}
\FOR{$i = 1$ to $m$}
    \STATE{kBlocks $\gets $ empty list}
    \FOR{$j = 1$ to $n$}
        \STATE{kBlock $\gets$ $\text{dropout}(R, rate=\widehat{R}(i, j))$}
        \STATE{kBlock $\gets$ shuffleColumnsAndRows(kBlock)}
        \STATE{kBlocks append kBlock}
    \ENDFOR
    \STATE{outputToFile(kBlocks)}
\ENDFOR
\end{algorithmic}
\end{algorithm}

\subsubsection{Preventing leaks from the test set into the training set}
For matrix factorization tasks, the usual procedure to build disjoint training and test data sets for Movie Lens consists in selecting some ratings and removing them from the training set while adding them to the test set.
A naive adaptation of the test data generation procedure to our extended data set would select test items directly on the larger matrix $\widetilde{R}$.
Unfortunately, as $\widetilde{R} = \widehat{R} \otimes R$, such a procedure would implicitly share data between interactions of the training and test sets through $\widehat{R}$ which incorporates information from the entire original data set $R$.
In order to generate training and test data without leaking test data into the training set, we proceed as follows.
We consider two separate training and test sets selected from the original data set $R$: $R_{\text{train}}$ and $R_{\text{test}}$.
With $R \in \mathbb{R}^{m, n}$, we have $R_{\text{train}} \in \mathbb{R}^{m, n}$ and $R_{\text{test}} \in \mathbb{R}^{m, n}$.
For the MovieLens data set, where each rating of a given item by a specific user is timestamped, a typical approach to defining training and testing sets removes the last rating of each user from the train set and adds it to the test set.
Such a procedure outputs a $R_{\text{test}}$ matrix with much fewer non zero elements than $R_{\text{train}}$. 

The smaller matrix $\widehat{R_{\text{train}}}$ is now derived from $R_{\text{train}}$ exclusively, without incorporating any data from $R_{\text{test}}$.
We create the extended versions of the train and test data sets separately as follows:
$$
    \widetilde{R_{\text{train}}} = \widehat{R_{\text{train}}} \odot R_{\text{train}}
    \text{ and }
    \widetilde{R_{\text{test}}} = \widehat{R_{\text{train}}} \odot R_{\text{test}}
    .
$$
By construction, the procedure prevents test data from leaking into the train data and implicitly informing the model of patterns that will be present in the test set during training.

\subsubsection{Consistent randomized operations across training and testing sets}
With a stochastic Kronecker $\odot_{\text{rand}}$ featuring dropout and block-wise shuffling, additional precautions need to be taken. In order to guarantee that the randomized shuffles of rows and columns are consistent between the training and testing data, we flip the sign of the test elements in the rating matrix to keep track of their belonging to the test set. We apply all randomized operations to the resulting matrix comprising elements in $\left\{-1, 0, 1\right\}$:
$$
    \widetilde{R_{\text{temp}}} = \widehat{R_{\text{train}}} \odot_{\text{rand}}
    \left(
        R_{\text{train}} - R_{\text{test}}
    \right)
$$
where $R_{\text{train}} \in \mathbb{R}^{m, n}$ and $R_{\text{test}} \in \mathbb{R}^{m, n}$.
The positive elements of $R_{\text{temp}}$ are attributed to $\widetilde{R_{\text{train}}}$ and the negative elements are attributed to $\widetilde{R_{\text{test}}}$ after having flipped their sign:
\begin{align*}
    & \forall i \in \left\{1, \dots, mp\right\}, j \in \left\{1, \dots, nq \right\}, \\
    & \widetilde{R_{\text{train}}}_{i,j} = \widetilde{R_{\text{temp}}}_{i,j} \text{ if } \widetilde{R_{\text{temp}}}_{i,j} = 1 \text{ else } 0\\
    & \widetilde{R_{\text{test}}}_{i,j} = - \widetilde{R_{\text{temp}}}_{i,j} \text{ if } \widetilde{R_{\text{temp}}}_{i,j} = -1 \text{ else } 0\\
\end{align*}

Such an operation is simple and guarantees the consistency of randomized shuffles of the sub-blocks in the extended matrices $\widetilde{R_{\text{train}}}$ and $\widetilde{R_{\text{test}}}$.

\section{Statistical properties of Kronecker fractal expansions}\label{sec:theory}
After having introduced Kronecker products to self-similarly expand a recommendation dataset into a much larger one, we now develop theoretical insights about how the transform preserves crucial common properties with the original.

\subsubsection{Salient empirical facts in MovieLens data}
First, we introduce the critical properties we want to preserve.
As a user/item interaction dataset on an online platform, one expects MovieLens to feature common properties of recommendation data sets such as power-law or fat-tailed distributions~\cite{zhao2018categorical}
(a power-law or fat-tailed distribution over positive values behaves like $x \rightarrow \alpha x^{-\beta}$ for large enough values of $x$ with $\alpha > 0$ and $\beta > 0$).

First important statistical properties for recommendations concern the distribution of interactions across users and across items. It is generally observed that such distributions exhibit a power-law behavior~\cite{zhao2018categorical,abdollahpouri2017controlling,oestreicher2012recommendation,cremonesi2010performance,park2008long,yin2012challenging,levy2010music,goel2010anatomy}.
To characterize such a behavior in the MovieLens data set, we take a look at the distribution of the total ratings along the item axis and the user axis. In other words, we compute row-wise and column-wise sums for the rating matrix $R$ and observe their distributions.
The corresponding ranked distributions are exposed in Figure~\ref{fig:original_properties} and do exhibit a clear power-law behavior for rather popular items. However, we observe that tail items have a higher popularity decay rate. Similarly, the engagement decay rate increases for the group of less engaged users.

The other approximate power-law we find in Figure~\ref{fig:original_properties} lies in the singular value spectrum of the MovieLens dataset.
We compute the top $k$ singular values~\cite{horn1990matrix} of the MovieLens rating matrix $R$ by power iteration, which can scale to its large $(138K, 27K)$ dimension.
The method yields the dominant singular values of $R$ and the corresponding singular vectors so that one can classically approximate $R$ by
$
    R \simeq U \Sigma V
$
where $\Sigma$ is diagonal of dimension $(k, k)$,
$U$ is column-orthogonal of dimension $(m, k)$ and $V$ is row-orthogonal of dimension $(k, n)$ ---
which yields the rank $k$ matrix closest to $R$ in Frobenius norm.

Examining the distribution of the $2048$ top singular values of $R$ in the MovieLens dataset (which has at most $27K$ non-zero singular values) in Figure~\ref{fig:original_properties} highlights a clear power-law behavior in the highest magnitude part of the spectrum of $R$.
We observe in the spectral distribution an inflection for smaller singular values whose magnitude decays at a higher rate than larger singular values.
Such a spectral distribution is as a key feature of the original dataset. This property is particularly important for low-rank approximation approaches to the matrix completion problem, which have to choose a sufficiently large rank for approximating the observations.
Therefore, we also want the expanded dataset to exhibit a similar behavior in terms of spectral properties. 

In all the high level statistics we present, we want to preserve the approximate power-law decay as well as its inflection for smaller values. Our requirements for the expanding transform which we apply to $R$ are therefore threefold: \emph{we want to preserve the distributions of row-wise sums of $R$, column-wise sums of $R$ and singular value distribution of $R$}.
Additional requirements, beyond first and second order high level statistics will further increase the confidence in the realism of the expanded synthetic dataset.

\subsubsection{Analytic tractability through standard Kronecker products}
Although we use a randomized version of the Kronecker product which does not offer the same level of analytic tractability, the choice of such a transform is deeply anchored in some of the theoretical properties of the standard Kronecker product.
We now expose how --- in its standard deterministic version --- the fractal transform design we rely on preserves the key statistical properties of the previous section.

\begin{definition}
Consider $A \in \mathbb{R}^{m, n} = (a_{i, j})_{i=1 \dots m, j=1 \dots n}$,
we denote the set $\left \{ \sum_{j = 1}^n a_{i, j}\right \}$ of row-wise sums of $A$  by $\mathcal{R}(A)$,
the set $\left \{ \sum_{i = 1}^m a_{i, j}\right \}$ of column-wise sums of $A$  by $\mathcal{C}(A)$,
and the set of non-zero singular values of $A$ by $\mathcal{S}(A)$.
\end{definition}

\begin{definition}
Consider an integer $i$ and a non-zero positive integer $p$,
we denote the integer part of $i - 1$ in base $p$
$
\lfloor{i - 1\rfloor}_p = \lfloor{\frac{i - 1}{p}\rfloor}
$
and the fractional part
$
\left\{i - 1\right\}_p = i - \lfloor{i - 1\rfloor}_p \times p.
$
\end{definition}

First we focus on conservation properties in terms of row-wise and column-wise sums which correspond respectively to marginalized user engagement and item popularity distributions.
In the following, $\times$ denotes the Minkowski product of two sets, i.e. 
$A \times B = \left\{ a \times b \: | \: \forall a \in A, \forall b \in B \right\}$.

\begin{proposition}
\label{prop:c_r_distributions}
Consider $A \in \mathbb{R}^{m, n}$ and $B \in \mathbb{R}^{p, q}$ and
their Kronecker product $K = A \otimes B$. Then
$$\mathcal{R}(K) = \mathcal{R}(A) \times \mathcal{R}(B) \text{ and }
\mathcal{C}(K) = \mathcal{C}(A) \times \mathcal{C}(B).$$
\end{proposition}

\begin{proof}
Consider the $i^{th}$ row of $K$, by definition of $K$ the corresponding sum can be rewritten as follows: 
$
\sum_{j = 1 \dots n p} k_{i, j} 
=
\sum_{j = 1 \dots n p} 
a_{\lfloor{i - 1\rfloor}_p + 1, \lfloor{j - 1\rfloor}_q + 1} 
b_{\left\{i - 1\right\}_p, \left\{j - 1\right\}_q}$
which in turn equals
$$
\sum_{j = 1 \dots n}
\sum_{j'= 1 \dots q}
a_{\lfloor{i - 1\rfloor}_p + 1, j}
b_{\left\{i - 1\right\}_p, j'}.
$$
Refactoring the two sums concludes the proof for the row-wise sum properties. The proof for column-wise properties is identical.
$\blacksquare$
\end{proof}

\begin{theorem}
\label{th:spectral_distribution}
Consider $A \in \mathbb{R}^{m, n}$ and $B \in \mathbb{R}^{p, q}$ and
their Kronecker product $K = A \otimes B$. Then
$$\mathcal{S}(K) = \mathcal{S}(A) \times \mathcal{S}(B).$$
\end{theorem}
\begin{proof}
One can easily check that $(X Y) \otimes (V W) = (X \otimes V) (Y \otimes W)$ for any quadruple of matrices $X,Y,V,W$ for which the notation makes sense and that $(X \otimes Y)^T = X^T \otimes Y^T$.
Let $A = U_A \Sigma_A V_A$ be the SVD of $B$ and $B = V_B \Sigma_B V_B$ the SVD of $B$.
Then $(A \otimes B) = (U_A \otimes U_B) (\Sigma_A \otimes \Sigma_B) (V_A \otimes V_B)$.
Now, $(U_A \otimes U_B)^T (U_A \otimes U_B) = (U_A^T \otimes U_B^T) (U_A \otimes U_B) = (U_A^T U_A) \otimes (U_B^T U_B^T)$.
Writing the same decomposition for $(V_A \otimes V_B) (V_A^T \otimes V_B^T)$ and considering that $U_A$, $U_B$ are column-orthogonal while $V_A$, $V_B$ are row-orthogonal concludes the proof.
$\blacksquare$
\end{proof}

The properties above imply that knowing the row-wise sums, column-wise sums and singular value spectrum of the reduced rating matrix $\widehat{R}$ and the original rating matrix $R$ is enough to deduce the corresponding properties for the expanded rating matrix $\widetilde{R}$ --- analytically. As in~\cite{leskovec2010kronecker}, the Kronecker product enables analytic tractability while expanding data sets in a fractal manner to orders of magnitude more data.

In practice, we use a randomized version of the Kronecker product whose block-wise shuffles do not have an analytically tractable effect of the high order statistics of the rating matrix.
Therefore, we rely in section~\ref{sec:ext_stats} on a statistical examination of the properties of the extended synthetic data set we produce with our randomized fractal operator to verify that our original theoretical insights from the deterministic case are still valid.
In particular, we demonstrate that original high order statistics of the new data set we produce preserve --- as in our first deterministic approach~\cite{belletti2019scalable} --- the original properties of the binary MovieLens 20m data set.

\subsubsection{Constructing a reduced $\widehat{R}$ matrix with a similar spectrum}
Considering that the quasi power-law properties of $R$ imply --- as in~\cite{leskovec2010kronecker} --- that $\mathcal{S}(R) \times \mathcal{S}(R)$ has a similar distribution to $\mathcal{S}(R)$, we seek a small $\widehat{R}$ whose high order statistical properties are similar to those of $R$.
As we want to generate a dataset with several billion user/item interactions, millions of distinct users and millions of distinct items, we are looking for a matrix $\widehat{R}$ with a few hundred or thousand rows and columns.
The reduced matrix $\widehat{R}$ we seek is therefore orders of magnitude smaller than $R$.
% \textbf{First attempts through randomized sketching:} 
% Random projections~\cite{achlioptas2003database,li2006very,fradkin2003experiments} appear as valid candidates to reduce the size of $R$ down while preserving its spectral properties.
% Indeed, proper random projections aim at finding $\widehat{R}$ such that
% $\widehat{R} \widehat{R}^T \simeq R R^T$\
% Such methods unfortunately do not produce $\widehat{R}$ with similar properties to $R$ as the dimensions of $\widehat{R}$ are too small compared to $R$. For random projections, such an issue can be interpreted as a consequence of having few dimensions in the bounds given by the Johnson-Lindenstrauss lemma. For random hashing, too few hashing buckets lead to high levels of aliasing which in turn skew the spectrum of the reduced matrix $\widehat{R}$.
In order to produce a reduced matrix $\widehat{R}$ of dimensions $(1000, 1700)$ one could use the reduced size older MovieLens 100K dataset~\cite{harper2016movielens}.
Such a dataset can be interpreted as a sub-sampled reduced version of MovieLens 20m with similar properties. These data sets have been collected seven years apart, wherein the characteristics of the dataset are not comparable. Also, we aim to produce an expansion method where the expansion multipliers can be chosen flexibly by practitioners. 
In our experiments, it is noteworthy that naive uniform user and item sampling strategies have not yielded smaller matrices $\widehat{R}$ with similar properties to $R$ in our experiments.
Different random projections~\cite{achlioptas2003database,li2006very,fradkin2003experiments} could more generally be employed. However, we rely on a procedure better tailored to our specific statistical requirements.

We now describe the technique we employed to produce a reduced size matrix $\widehat{R}$ with first and second order properties close to $R$ which in turn led to constructing an expansion matrix $\widetilde{R} = \widehat{R} \otimes R$ similar to $R$.
We want the dimensions of $\widehat{R}$ to be $(m',n')$ with $m' << m$ and $n' << n$.
Consider again the approximate Singular Value Decomposition (SVD)~\cite{horn1990matrix} of $R$ with the $k=min(m',n')$ principal singular values of $R$:
\begin{equation}
    \label{eq:R_svd}
    R \simeq U \Sigma V
\end{equation}
where $U \in \mathbb{R}^{n, k}$ has orthogonal columns, 
$V \in \mathbb{R}^{k, m}$ has orthogonal rows, 
and $\Sigma \in \mathbb{R}^{k, k}$ is diagonal with non-negative terms.

To reduce the number of rows and columns of $R$ while preserving its top $k$ singular values a trivial solution would consist in replacing $U$ and $V$ by a small random orthogonal matrices with few rows and columns respectively.
Unfortunately such a method would only seemingly preserve the spectral properties of $R$ as the principal singular vectors would be widely changed. Such properties are important: one of the key advantages of employing Kronecker products in~\cite{leskovec2010kronecker} is the preservation of the network values, i.e. the distributions of singular vector components of a graph's adjacency matrix.

To obtain a matrix $\widetilde{U} \in \mathbb{R}^{n', k}$ with fewer rows than $U$ but column-orthogonal and similar to $U$ in the distribution of its values we use the following procedure.
We re-size $U$ down to $n'$ rows with $n' < n$ by down-scaling through local averaging (using skimage.transform.resize in the scikit-image library~\cite{van2014scikit}).
Let $\bar{U} \in \mathbb{R}^{n', k}$ be the corresponding resized version of $U$. 
We then construct $\widetilde{U}$ as the column orthogonal matrix in $\mathbb{R}^{n', k}$ closest in Frobenius norm to $\bar{U}$.
Therefore as in~\cite{golub2012matrix} we compute
\begin{equation}
    \label{eq:U_reduction}
    \widetilde{U} = \bar{U} \left(\bar{U}^T \bar{U}\right)^{-1/2}.  
\end{equation}
We apply a similar procedure to $V$ to reduce its number of columns which yields a row orthogonal matrix $\widetilde{V} \in \mathbb{R}^{k, m'}$ with $m' < m$.
The orthogonality of $\widetilde{U}$ (column-wise) and $\widetilde{V}$ (row-wise) guarantees that the singular value spectrum of
\begin{equation}
    \label{eq:R_hat_construction}
    \widehat{R} = \widetilde{U} \Sigma \widetilde{V}
\end{equation}
consists exactly of the $k=min(m',n')$ leading components of the singular value spectrum of $R$.
Like $R$, $\widehat{R}$ is re-scaled to take values in $[-1, 1]$.
The whole procedure to reduce $R$ down to $\widehat{R}$ is summarized in Algorithm~\ref{alg:reduced_matrix}.

\begin{algorithm}
\caption{Compute reduced matrix $\widehat{R}$}
\label{alg:reduced_matrix}
\begin{algorithmic}
\STATE{$(U, \Sigma, V) \gets$ sparseSVD($R,k$)}
\STATE{$\bar{U} \gets$ imageResize($U, n', k$)}
\STATE{$\bar{V} \gets$ imageResize($V, k, m'$)}
\STATE{$\widetilde{U} \gets \bar{U} \left(\bar{U}^T \bar{U}\right)^{-1/2}$}
\STATE{$\widetilde{V} \gets \left(\bar{V} \bar{V}^T\right)^{-1/2} \bar{V}$}
\STATE{$\widehat{R}_{temp} \gets \widetilde{U} \Sigma \widetilde{V}$}
\STATE{$M \gets max(\widehat{R}_{temp})$}
\STATE{$m \gets min(\widehat{R}_{temp})$}
\STATE{return $\widehat{R}_{temp} / (M - m)$}
\end{algorithmic}
\end{algorithm}

We verify empirically that the distributions of values of the reduced singular vectors in $\widetilde{U}$ and $\widetilde{V}$ are similar to those of $U$ and $V$ respectively to preserve first order properties of $R$ and value distributions of its singular vectors.
Such properties are demonstrated through numerical experiments in the next section.

\section{Experimentation on MovieLens 20 million data}
The MovieLens $20M$ data comprises $20M$ ratings given by $138K$  users to $27K$  items.
In the present section, we demonstrate how the fractal Kronecker expansion technique we devised and presented helps scale up this dataset to orders of magnitude more users, items and interactions --- all in a parallelizable manner.

\subsubsection{Pre-processing of MovieLens 20m}
The first pre-processing step we apply to MovieLens 20m is binarizing all the ratings: all the rating values are set to $1$.
Such a step is standard for tasks such as Neural Collaborative Filtering~\cite{he2017neural}.
The second pre-processing step filters out users who have fewer than $2$ ratings with distinct timestamps.
The filter enables the splitting of MovieLens 20m into a train set $R_{\text{train}}$ consisting of all the ratings of each users except the last one in chronological order and its complement. For each user, the rating with the latest timestamp is put in the test set $R_{\text{test}}$.
After removal of users with too few ratings and splitting into training and test sets, we expand MovieLens 20m with the randomized Kronecker product presented in Algorithm~\ref{alg:kron_randomized_exp}.

\subsubsection{Size of expanded data set}

In the present experiments we construct a reduced rating matrix $\widehat{R}$ of size $(16, 32)$.
The dropout based method in Algorithm~\ref{alg:kron_randomized_exp}
yields a new data set whose size is detailed in Table~\ref{tab:extended_size}.

\begin{table}[]
    \centering
    \begin{tabular}{c|c|c|c}
        & MovieLens 20m & Synthetic train set & Synthetic test set \\
        \hline
        Interactions & $20$ M & $1.22$ B & $12.7$ M\\
        \hline
        Users & $138$ K & $2.20$ M & $2.20$ M\\
        \hline
        Items & $27$ K & $855$ K & $855$ K\\
        \hline
    \end{tabular}
    \caption{Size of the extended MovieLens20m data set}
    \label{tab:extended_size}
\end{table}

Such a high number of interactions and items enable the training of deep neural collaborative models such as the Neural Collaborative Filtering model~\cite{he2017neural} with a scale which is now more representative of industrial settings.
Moreover, the increased data set size helps construct benchmarks for deep learning software packages and ML accelerators that employ the same orders of magnitude as production settings in terms of user base size, item vocabulary size and number of observations.

\subsubsection{Empirical properties of reduced $\widehat{R}$ matrix}

The objective of the construction technique for $\widehat{R}$ was to produce a matrix sharing the properties of $R \otimes R$ though smaller in size (~\cite{leskovec2010kronecker}).
To that end, we aimed at constructing a matrix $\widehat{R}$ of dimension $(16, 32)$ with properties close to those of $R$ in terms of column-wise sum, row-wise sum and singular value spectrum distributions.

We now check that the construction procedure we devised does produce a $\widehat{R}$ with the properties we expected.
As the impact of the re-sizing step is unclear from an analytic stand-point, we had to resort to numerical experiments to validate our method.

\begin{figure}
    \centering
    \includegraphics[width=\linewidth]{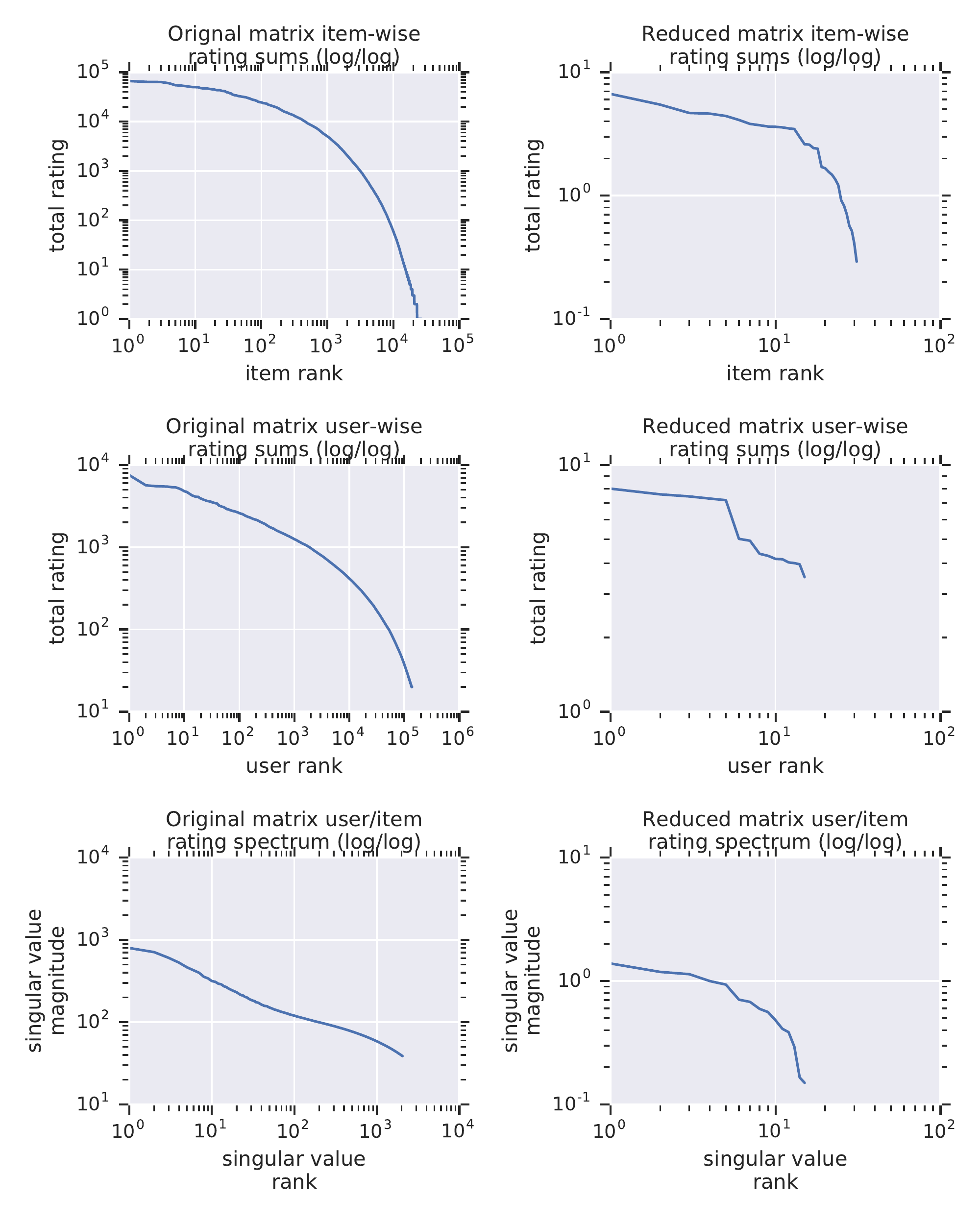}
    \caption{Properties of the reduced dataset $\widehat{R} \in \mathbb{R}^{16, 32}$ built according to steps~\ref{eq:R_svd}, \ref{eq:U_reduction} and \ref{eq:R_hat_construction}. We validate the construction method numerically by checking that the distribution of row-wise sums, column-wise sums and singular values are similar between $R$ and $\widehat{R}$. Note here that as $R$ is large we only compute its leading singular values. As we want to preserve statistical ``power-laws'', we focus on preservation of the relative distribution of values and not their magnitude in absolute.}
    \label{fig:small_properties}
\end{figure}

In Figure~\ref{fig:small_properties}, one can assess that the first and second order properties of $R$ and $\widehat{R}$ match with high enough fidelity. In particular, the higher magnitude column-wise and row-wise sum distributions follow a ``power-law'' behavior similar to that of the original matrix. Similar observations can be made about the singular value spectra of $\widehat{R}$ and $R$.

There is therefore now a reasonable likelihood that our adapted Kronecker expansion --- although somewhat differing from the method originally presented in~\cite{leskovec2010kronecker} --- will enjoy the same benefits in terms of enabling data set expansion while preserving high order statistical properties.

\subsubsection{Empirical properties of the expanded data set $\widetilde{R}$\label{sec:ext_stats}}
We now verify empirically that the expanded rating matrix $\widetilde{R} = \widehat{R} \otimes R$ does share common first and second order properties with the original rating matrix $R$. The new data size is $2$ orders of magnitude larger in terms of number of rows and columns and $4$ orders of magnitude larger in terms of number of non-zero terms.
Notice here that because of the dropout, the density of the resulting data set is about$~20\%$ that of the original data set.

\begin{figure}
    \centering
    \includegraphics[width=\linewidth]{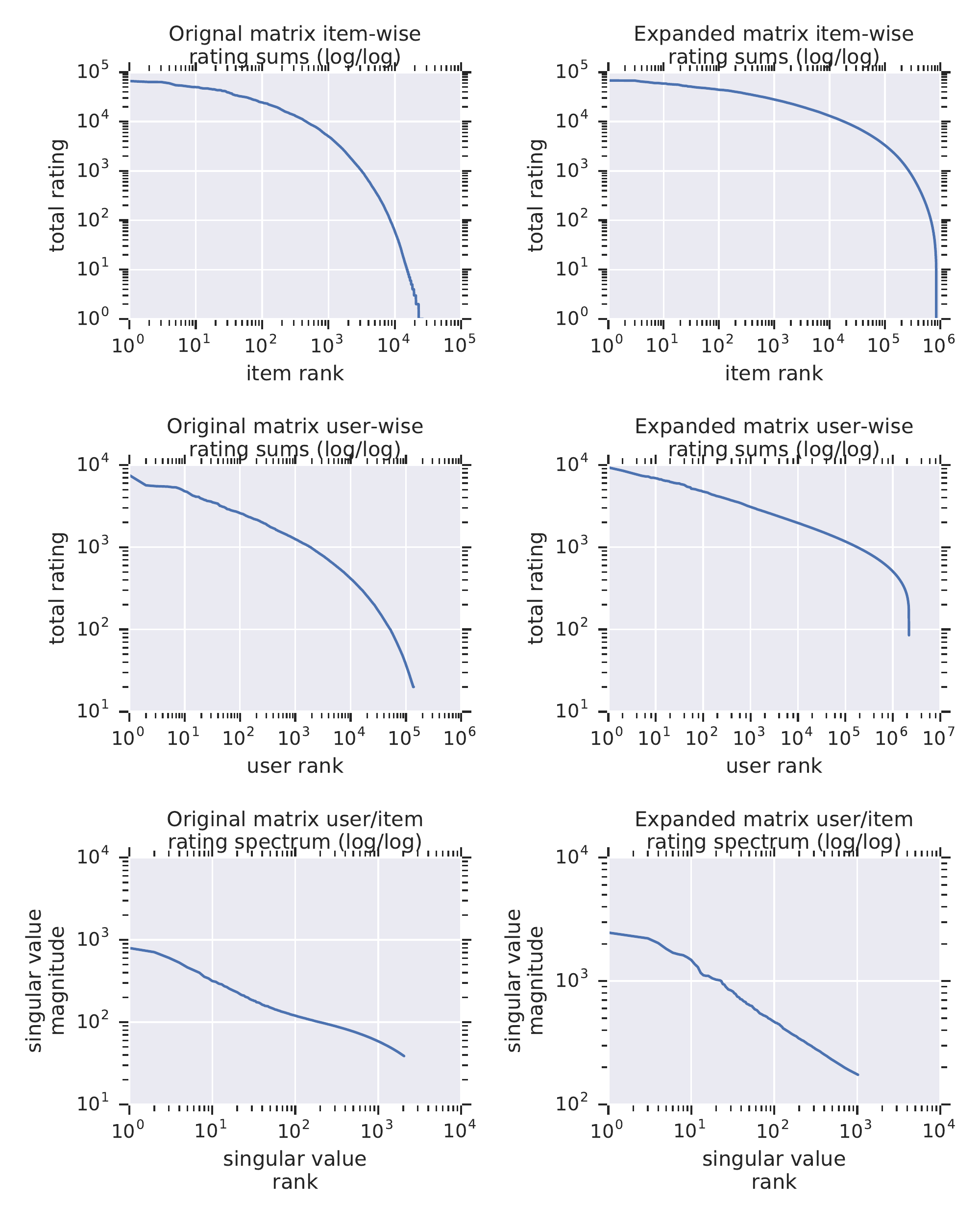}
    \caption{High order statistical properties of the expanded dataset $\widehat{R} \otimes R$. We validate the construction method numerically by checking that the distributions of row-wise sums, column-wise sums and singular values are similar between $R$ and $\widehat{R} \otimes R$. Here we leverage the tractability of Kronecker products as they impact column-wise and row-wise sum distributions as well as singular value spectra.
    In all plots we can observe the preservation of the linear log-log correspondence for the higher values in the distributions of interest (row-wise sums, column-wise sums and singular values) as well as the accelerated decay of the smaller values in those distributions.
    }
    \label{fig:expanded_properties}
\end{figure}

In Figure~\ref{fig:expanded_properties}, one can confirm that the spectral properties of the expanded data set as well as the user engagement (row-wise sums) and item popularity (column-wise sums) are similar to those of the original data set.
Such observations demonstrate that the theoretical insights from Proposition~\ref{prop:c_r_distributions} and Theorem~\ref{th:spectral_distribution} are indeed informative of the high order statistics of the synthetic data set we generate.
Our ex-post empirical study indicates that the resulting data set is representative --- in its fat-tailed data distribution and quasi ``power-law'' singular value spectrum --- of problems encountered in ML for collaborative filtering.
Furthermore, the expanded data set reproduces some unique properties of the original data, in particular the accelerating decay of values in ranked row-wise and column-wise sums as well as in the singular values spectrum.

\section{Conclusion}
In conclusion, this paper presents an attempt at synthesizing a realistic large-scale recommendation data sets without having to make compromises in terms of user privacy.
We use a small size publicly available data set, MovieLens 20m, and expand it to orders of magnitude more users, items and observed ratings.
Our expansion model is rooted into the hierarchical structure of user/item interactions which naturally suggests a fractal extrapolation model.

We leverage randomized Kronecker products as self-similar operators on user/item rating matrices that preserve key properties of row-wise and column-wise sums as well as singular value spectra.
We modify the original Kronecker Graph generation method to enable a randomized expansion of the original data by orders of magnitude that yields a synthetic data set matching industrial recommendation data sets in scale.
Our numerical experiments demonstrate the data set we create has key first and second order properties similar to those of the original MovieLens 20m binarized rating matrix.

Our next steps consist in making large synthetic data sets publicly available although any researcher can readily use the techniques we presented to scale up any user/item interaction matrix.
Another possible direction is to adapt the present method to recommendation data sets featuring metadata (e.g. timestamps, topics, device information).
The use of metadata is indeed critical to solve the ``cold-start'' problem of users and items having no interaction history with the platform.
In this work, we did not consider here the temporal structure of the MovieLens data set. We leave the study of sequential user behavior --- often found to be Long Range Dependent~\cite{pipiras2017long,tang2019towards,belletti2017random,crane2008robust} --- and the extension of synthetic data generation to sequential recommendations~\cite{belletti2018factorized,tang2018caser,yu2016dynamic} for further work.
We also plan to benchmark the performance of well established baselines on the new large scale realistic synthetic data we produce.

\bibliographystyle{acm}
{\small
\bibliography{main}
}
\newpage

\end{document}